\documentclass[10pt,twocolumn,letterpaper]{article}

\usepackage{titling}
\usepackage{cvpr}
\usepackage{times}
\usepackage{epsfig}
\usepackage{graphicx}
\usepackage{amsmath}
\usepackage{amssymb}
\usepackage{authblk}
\usepackage{siunitx}
\usepackage{subfigure}
\usepackage{pifont}

\usepackage[breaklinks=true,bookmarks=false]{hyperref}
\usepackage{cleveref}

\cvprfinalcopy
\def\cvprPaperID{8428}

\ifcvprfinal\fi

\makeatletter
\newcommand{\emptyauthor}{%
\renewcommand{\author}[1]{}
\renewcommand{\@author}{}}
\makeatother

\begin{document}

\title{Digital Gimbal: End-to-end Deep Image Stabilization with Learnable Exposure Times}
\author{Omer Dahary}
\author{Matan Jacoby}
\author{Alex M. Bronstein}
\affil[1]{Technion - Israel Institute of Technology}
\affil[ ]{{\tt\small\{omerd,matanj,bron\}@cs.technion.ac.il}}
\date{}
\maketitle

\begin{abstract}
    Mechanical image stabilization using actuated gimbals enables capturing long-exposure shots without suffering from blur due to camera motion. These devices, however, are often physically cumbersome and expensive, limiting their widespread use. In this work, we propose to digitally emulate a mechanically stabilized system from the input of a fast unstabilized camera. To exploit the trade-off between motion blur at long exposures and low SNR at short exposures, we train a CNN that estimates a sharp high-SNR image by aggregating a burst of noisy short-exposure frames, related by unknown motion. We further suggest learning the burst's exposure times in an end-to-end manner, thus balancing the noise and blur across the frames. We demonstrate this method's advantage over the traditional approach of deblurring a single image or denoising a fixed-exposure burst on both synthetic and real data.
\end{abstract}

\thispagestyle{empty}

\section{Introduction}

\begin{figure}[h]
\begin{center}
\subfigure[Full-exposure image]
{
    \includegraphics[width=0.47\linewidth]{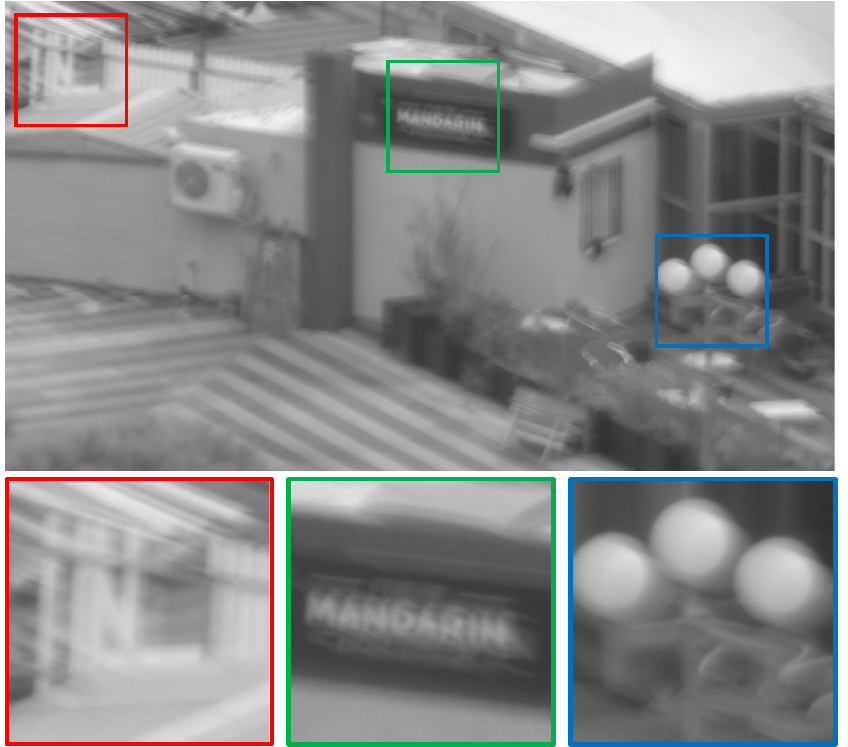}
}
\subfigure[DeblurGAN-v2~\cite{kupyn2019deblurgan}]
{
    \includegraphics[width=0.47\linewidth]{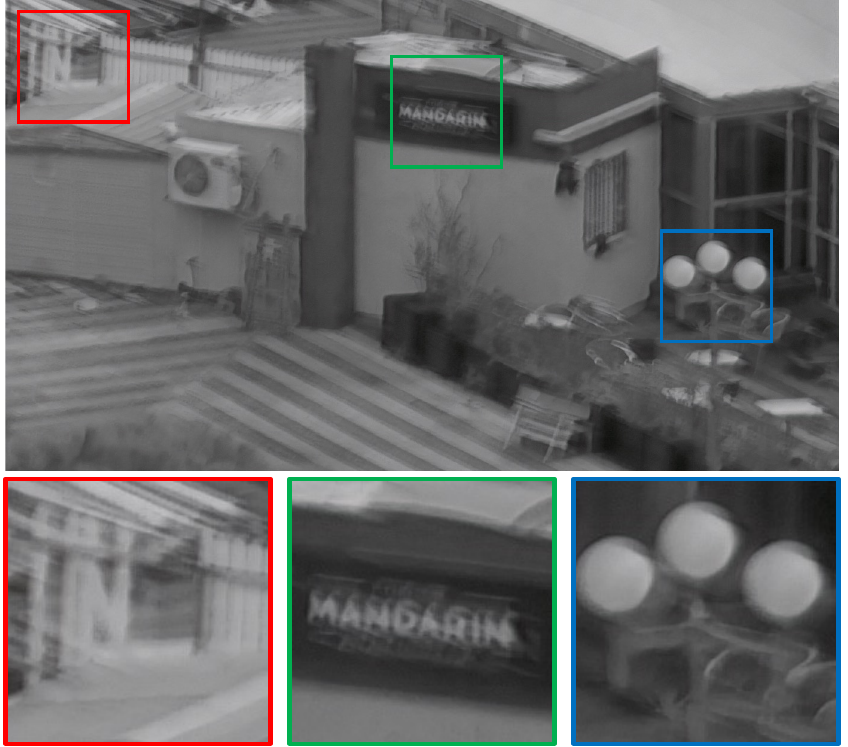}
}
\\
\subfigure[KPN~\cite{mildenhall2018burst}]
{
    \includegraphics[width=0.47\linewidth]{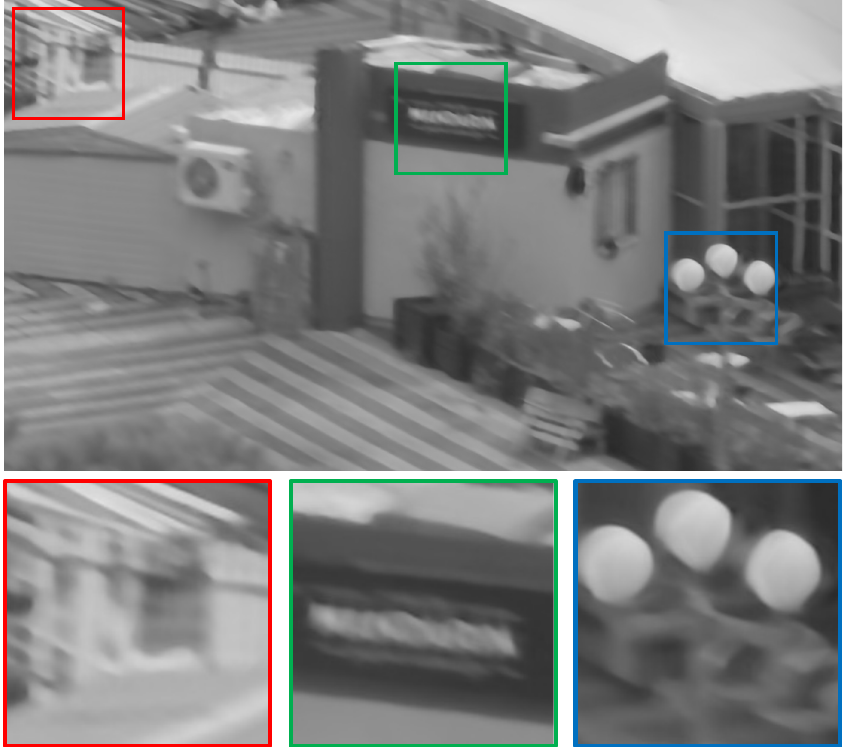}
}
\subfigure[Ours]
{
    \includegraphics[width=0.47\linewidth]{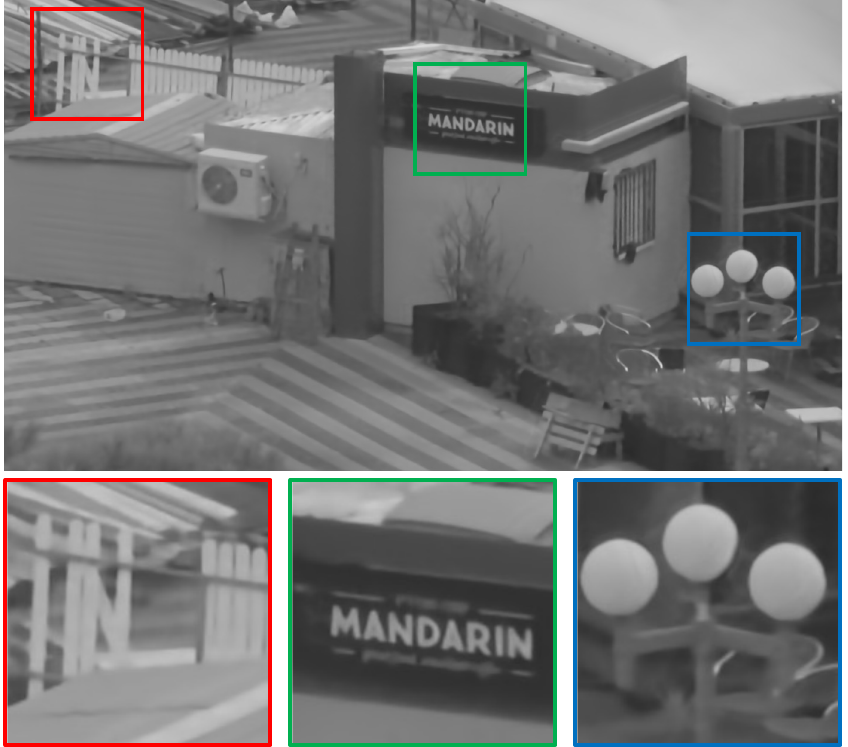}
}
\end{center}
   \caption{A qualitative evaluation of our approach on images captured by a vibrating camera with a long focal length. (See the supplement for a description of our setup.) The fully-exposed image (a) suffers from considerable motion blur, which is hard to mitigate using image deblurring methods (b). Employing the same time budget to capture a fixed-exposure burst and merging it via a burst denoising algorithm (c) results in over-smoothing and ghosting artifacts. Our approach (d) of learning a non-uniform exposure regime enables utilizing the sharpness of short-exposed frames and the high SNR of long-exposed ones to yield a superior reconstruction.}
\label{fig:real_out}
\end{figure}

Through advances in imaging technology, optical systems have become lighter and more portable than ever. These improvements, however, are sometimes negated by the need for stabilization. Due to the natural tremor of handheld cameras, or the movement of mounted vehicles, camera motion is often unavoidable. As a result, images captured this way may suffer from motion blur, which is especially evident during long exposures or with long focal lengths. Image stabilization, the task of mitigating this phenomenon, has been mostly explored in the optical and mechanical domains. In the optical regime, stabilization emerges from using high-cost, uniquely tailored lenses or cheaper but less effective shift sensors. On the other hand, mechanical stabilization devices, such as actuated gimbals, can be attached to any camera with no need for complicated optical equipment. Such systems are highly popular but limited in their ability to compensate for blur due to scene motion or atmospheric turbulence. However, the most severe limitation of mechanical image stabilization is its cost, form factor, weight, and power requirements, which might be prohibitive in many scenarios.

In recent years, deep learning approaches have been successfully utilized for solving traditional image processing tasks. The field of burst and multi-frame imaging, in particular, has seen rising attention, with a variety of works on denoising~\cite{mildenhall2018burst} and deblurring~\cite{aittala2018burst,wieschollek2016end}. These methods exploit the information entangled in different degraded realizations of the same scene to reassemble them into a single high-quality image. Relying on this concept, together with the well-known trade-off between strong blur at long exposures and low SNR at short exposures, we propose solving the image stabilization task using a burst of images, each captured consecutively over a short period. Thus we reduce the problem to joint burst denoising and deblurring, as the two are already well-established. While most burst methods assume the data across the burst are complementary, we further suggest explicitly intervening in the acquisition pipeline and forcing diversity in noise and blur levels. We achieve this by designing an end-to-end learned system that includes both a deep burst image processing network and the camera's exposure configuration. The latter is implemented using a novel network layer that numerically models the frame acquisition process, and can potentially be further tuned for other imaging tasks. To the best of our knowledge, this approach has never been attempted before for still image reconstruction from SNR-limited bursts in the presence of camera motion.

\section{Related work}

Burst imaging aims to compensate for non-optimal optics or imaging conditions by fusing frames captured sequentially over a short period. This ongoing research field has been excessively studied for denoising and deblurring, with most works focusing on a single kind of degradation.

Delbracio and Sapiro~\cite{delbracio2015burst} propose an elegant scheme for weighted spectral domain averaging of uniformly-blurred bursts to produce sharp images. Wieschollek \etal~\cite{wieschollek2016end} expand on this idea and introduce a CNN for predicting deconvolution filters and weights for Fourier burst accumulation. Aittala and Durand~\cite{aittala2018burst} suggest a permutation-invariant network for fusing unordered sets of blurry images. Mildenhall \etal~\cite{mildenhall2018burst} propose an adaptive kernel prediction network for jointly aligning and merging noisy frames. Sim and Kim~\cite{sim2019deep} further utilize this approach for video deblurring by fusing adjacent frames and a generated residual image.

The task of image restoration from bursts with non-uniform exposure times was first explored using traditional image processing tools. Ben-Ezra and Nayar~\cite{ben2003motion} introduce a hybrid imaging system that records camera motion to predict the point spread function of a blurry image. Zhang \etal~\cite{zhang2013multi} use a sparse prior to adaptively combine information from noisy and blurry inputs. Yuan \etal~\cite{yuan2007image} employ a noisy/blurry image pair of the same scene to estimate the blur kernel and reduce ringing artifact. More recently, researchers have been adopting the idea of using short- and long- exposure pairs as the input to deep learning-based restoration algorithms. These include tasks such as high dynamic range (HDR) imaging~\cite{kalantari2019deep}, deblurring~\cite{zhang2019deep}, and low-light restoration~\cite{chang2020low}. However, these works rely on fixed and pre-selected timing regimes, with two separately-tuned data generation pipelines: one for the short-exposure frames, and another one for the long-exposure ones.

In parallel, recent works by Google Research have shown that burst imaging can mitigate smartphone cameras' limitations by increasing their dynamic range~\cite{hasinoff2016burst}, spatial resolution~\cite{wronski2019handheld}, and performance in low light conditions~\cite{liba2019handheld}. These improvements are achieved by an alternative image signal processing (ISP) pipeline that robustly aligns and merges frames and applies post-processing effects. Furthermore, it introduces dynamic exposure selection: either by interpolating a hand-crafted database, or by aggregating motion prediction with inertial sensor data.

\section{Approach}
\label{sec:approach}

\begin{figure*}[h]
\begin{center}
\includegraphics[width=1\linewidth]{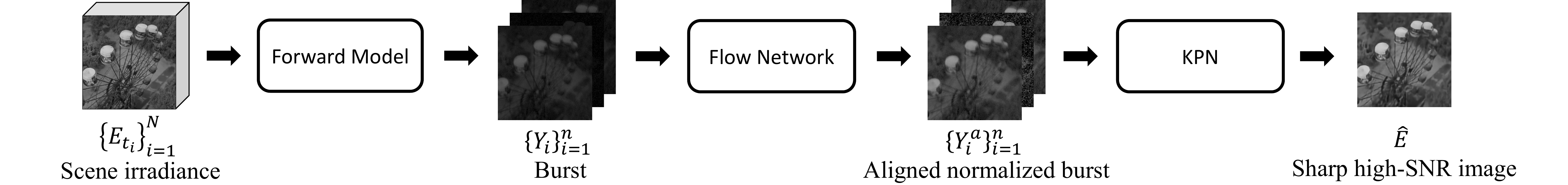}
\end{center}
   \caption{Our reconstruction network comprises a flow network, which aligns the burst, and a KPN, which merges the aligned frames. At train time, our forward model generates the burst from a video representing the scene's irradiance values. This extension allows learning the frames' optimal exposure times.}
\label{fig:scheme}
\end{figure*}

We assume the scenario in which a latent scene is captured by a moving camera over a temporal interval $[0,T]$. With some abuse, we denote the irradiance image at time $t$ as a latent transformation $\tau_t$ of a latent irradiance $E$, $E_t = \tau_t (E)$. The irradiance images are not directly observable; instead, the camera acquires a discrete set of $n$ frames $Y_1,\dots, Y_n$. Each frame is captured during its integration interval $\left[t_i,t_i + \Delta t_i\right] \subset \left[0,T\right]$,
where $t_i$ denotes the opening time of the shutter, and $\Delta t_i$ the frame exposure.
Each frame is related to the latent image through the sensor forward model $F$:
\begin{align}
    Y_i = F\left( \tau_{t \in [t_i, t_i+\Delta t_i] } (E)  \right).
\end{align}
Assuming the exposure times $\Delta t_i$ are relatively low (around a few milliseconds), each frame in the burst suffers from low SNR due to imaging noise. Furthermore, we expect the frames to be slightly blurred due to the camera movement. The trade-off between these two sources of degradation is controlled by the exposure time $\Delta t_i$, with longer exposures corresponding to better SNR and stronger blur, and vice versa. 

Our aim is to utilize the information entangled in these measurements of the scene, by combining the frames into a single sharp high-SNR estimation of the latent irradiance image,
\begin{align}
    \hat{E} = I\left(Y_1,\dots, Y_n\right),
\end{align}
where $I$ is the reconstruction (inverse) operator.

We propose to learn the latter reconstruction operator concurrently with the user-controlled parameters of the camera, which in our case is the shutter schedule $\{t_i, \Delta t_i\}$. This leads to a learning problem of the form
\begin{align}
    \min_{\mathbf{\Delta}, I} \mathbb{E} \ell\left(
    I\left(\left\{ F\left( \tau_{t \in [t_i, t_i+\Delta t_i] }(E) \right) \right\}_{i=1}^n \right), E
    \right).
\end{align}
Here $\mathbb{E} \ell$ denotes the expectation of a loss function measuring the discrepancy between the latent $E$ and its estimated version $\hat{E}$, and $\bf{\Delta}$ describes the burst parameters detailed in the sequel.
In what follows, we describe the construction of differentiable forward and inverse models and their learning scheme.

\section{Sensor forward model}

This section describes our numerical forward model (denoted as $F$ above) of the imaging process, starting with the scene irradiance and ending with a fully formed raw image. We generally follow the noise estimation of Konnik and Welsh~\cite{konnik2014high}, while paying particular attention to time-dependent properties and visibly dominant factors at short exposures.

\subsection{Image formation}

Our model describes a moving camera capturing short-exposure images of a static scene. For simplicity, we assume that both the camera and the light source are monochromatic in relation to the same wavelength $\lambda$. We further assume device-specific parameters are known, as they are customarily supplied by manufacturers~\cite{jahne2010emva}.

\paragraph{From photons to electrons.}

Let $E_t\left(\bf{x}\right)$ be the irradiance of a pixel $\bf{x}$ at time $t$. We can express it in terms of the photon flux
\begin{align}
    \gamma_t\left(\mathbf{x}\right) = \frac{\lambda A E_t\left(\mathbf{x}\right)}{h c},
    \label{eq:irradiance}
\end{align}
where $\lambda$ denotes the wavelength, $A$ is the effective pixel area, $h$ is the Planck constant, and $c$ is the speed of light.

Each collected photon deposits its energy in the form of electric charge, generating photoelectrons inside the pixel. Their mean amount is given by
\begin{align}
    \bar{e}(\mathbf{x}) = \eta_\lambda \int_{t}^{t + \Delta t} \gamma_t\left(\mathbf{x}\right) dt,
    \label{eq:integral}
\end{align}
where $[t,t + \Delta t]$ is the time interval on which the shutter was open, and $\eta_\lambda$ is the quantum efficiency of the pixel at wavelength $\lambda$.

\paragraph{Noise generation.}

The photons are not the sole source of charge in the pixel. Some amount of current, named the dark current, thermally deposits charge in it, even if the scene is completely dark. The mean number of electrons it generates grows linearly with the exposure time:
\begin{align}
    \bar{e}_0 = \frac{I_0 \Delta t}{q_e},
\end{align}
where $I_0$ is the average dark current, and $q_e$ is the elementary charge.

The measured number of electrons fluctuates randomly, obeying Poisson statistics due their discrete nature. This phenomenon, known as shot noise, is often approximated in the image processing domain by a Gaussian distribution~\cite{chang2020low,kalantari2019deep,mildenhall2018burst}, which is a valid assumption in high-light regimes. However, since we are dealing with short exposures, at which shot noise dominates other components, we model photoelectron production as a realization of a Poisson variable,
\begin{align}
    e_\mathrm{q}(\mathbf{x}) \sim \mathrm{Poiss}\left(\bar{e}(\mathbf{x}) +\bar{e}_0\right).
    \label{eq:shot}
\end{align}

Another major source of noise, the readout noise, is independent of the exposure time. It mostly originates in thermal fluctuations in the analog-to-digital converter (ADC) and follows a zero-mean Gaussian distribution with some device-specific standard deviation,
$e_\mathrm{ro}(\mathbf{x}) \sim \mathcal{N}\left(0,\sigma^2_\mathrm{ro}\right)$.

Therefore, we can express the total number of collected electrons as
\begin{align}
    e(\mathbf{x}) = e_\mathrm{q}(\mathbf{x}) + e_\mathrm{ro}(\mathbf{x}).
    \label{eq:total}
\end{align}

\paragraph{From electrons to digital numbers.}

The pixel circuits convert the collected electrons into voltage, which is then amplified and translated into digital numbers (DNs). Following the EMVA 1288 standard~\cite{jahne2010emva}, we assume this process to be described by an almost linear curve, whose sensitivity dampens towards the full-well capacity (FWC) of the pixel. We model this response function as
\begin{align}
    r\left(e\right)=
    \begin{cases}
        K e, & e\leq\tau_1 \\
        K\left(\tau_1 + \left(1-\exp\left(-\frac{e-\tau_1}{\tau_2}\right)\right)\tau_2\right), & e\geq\tau_1
    \end{cases},
\end{align}
where $K$ is the overall system gain, $\tau_1$ is the threshold from which the curve becomes strictly concave, and $\tau_1+\tau_2=\mathrm{FWC}$.
Moreover, since DNs have discrete values, we assume the following quantization function:
\begin{align}
    Q\left(e\right)=\min\left\{ \max\left\{ \frac{[e]}{2^m-1}, 0 \right\}, 1 \right\},
    \label{eq:quantization}
\end{align}
with $m$ denoting the number of bits per pixel, and $[\cdot]$ the rounding  operation.
The DN value of the digital image at pixel $\mathbf{x}$ is given by
\begin{align}
    Y(\mathbf{x}) = Q\left(r\left(e(\mathbf{x})-\bar{e}_0\right)\right),
    \label{eq:dn}
\end{align}
where the subtraction of the mean of the dark noise $\bar{e}_0$ reflects the typical correction that many cameras apply.

\subsection{Exposure parametrization}

\begin{figure}
\begin{center}
\includegraphics[width=1\linewidth]{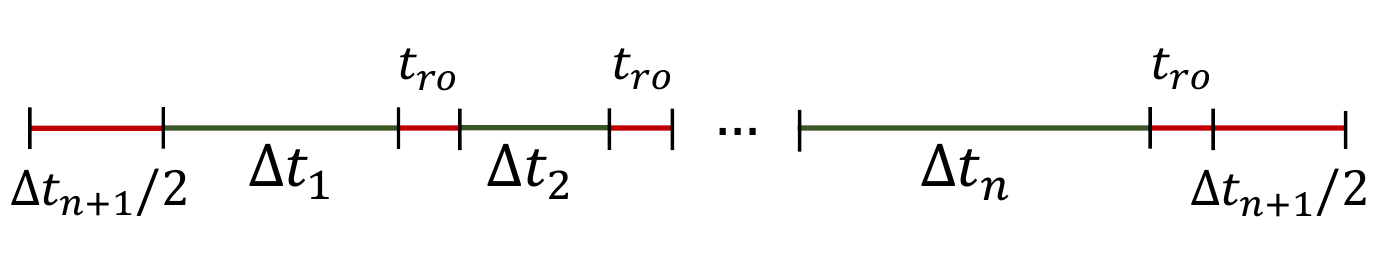}
\end{center}
   \caption{The shutter schedule is determined by the learned exposure times. Green and red lines represent time intervals in which the shutter is open and close, respectively. The idle time slot is optional.}
\label{fig:time}
\end{figure}

In order to make the sensor forward model amenable to learning, we assume the total time budget $T$ and the number of frames $n$ in the burst to be fixed, and parametrize the individual frame exposure parameters as 
\begin{align}
    \Delta t_i = \Delta t_{\mathrm{min}} + \alpha_i \left(T - n\left(\Delta t_{\mathrm{min}} + \Delta t_{\mathrm{ro}}\right)\right).
\end{align}
Here $\Delta t_{\mathrm{min}}$ is the minimum exposure time, and $t_\mathrm{ro}$ is the frame readout time (including any additional blank time needed between two consecutive frames). The parameters $\alpha_i$, representing the relative frame exposures, are, in turn, parametrized as
\begin{align}
    \alpha_i = \sigma\left(\bf{\Delta} \right)_i = \frac{e^{\Delta_i}}{\sum_{j=1}^m e^{\Delta_j}},
\end{align}
where $\sigma$ denotes softmax. The vector $\bf{\Delta}$, serving as the trainable camera parameters, can be set to be either $n$-dimensional, in which case the burst consumes the entire available time budget, or $(n+1)$-dimensional, allowing not to utilize all the available time (Fig. \ref{fig:time}).

\subsection{Numerical approximation}

The learning of the optimal burst exposure parameters $\bf{\Delta}$ mandates a differentiable calculation of the forward model described by equations (\ref{eq:integral}-\ref{eq:dn}). While most of these computations are straightforward, special consideration should be taken in steps (\ref{eq:integral},\ref{eq:shot},\ref{eq:quantization}).

The forward pass of (\ref{eq:integral}) calculates a temporal integral of a continuous irradiance function $E_t(\mathbf{x})$, scaled according to (\ref{eq:irradiance}). At training, we approximate this integral via the trapezoidal method from a sequence of $N \gg n$ uniformly-sampled irradiance values $\{ E_{t_i} (\mathbf{x}) \}_{i=1}^N$ obtained from simulated scene flow.

We discuss the backpropagation through equations (\ref{eq:shot},\ref{eq:quantization}) in the supplement.

\section{Reconstruction network}

The following section describes our inverse model (denoted as $I$ in Section~\ref{sec:approach}) responsible for estimating the clean irradiance image $E$ given the burst frames $Y_1,\dots,Y_n$.
We base our reconstruction model on the recent work in kernel prediction networks (KPN), which were first introduced by Jia \etal~\cite{jia2016dynamic}, and have shown promising results in various image processing tasks, including video interpolation~\cite{niklaus2017video1,niklaus2017video2}, super-resolution~\cite{jo2018deep}, and deblurring~\cite{sim2019deep}, and burst denoising~\cite{mildenhall2018burst}. Its success in the latter two problems encourages its use in this work, as our goal is to merge different noisy and blurry realizations of the same scene. To do so, we predict temporally- and spatially-variant kernels and apply them to merge the captured burst into a clean and sharp image.

Since fixed-size kernels are limited in their ability to align frames, we pre-warp the burst according to a reference one, as customarily applied in video deblurring works~\cite{kim2018spatio,su2017deep}. Gast and Roth~\cite{gast2019deep} proposed using pre-trained flow networks as the most efficient option for the latter operation. However, this approach is impractical in our case, as the amount of noise and blur in the input frames is expected to change while the exposure times are learned. Therefore, we opted for an end-to-end training methodology, in which the flow network parameters are co-learned, not specifically for perfect alignment, but to compensate for the kernels' narrow receptive field.

A schematic representation of our network is summarized in Fig. \ref{fig:scheme}.

\subsection{Flow network}

\begin{figure}
\begin{center}
\includegraphics[width=0.95\linewidth]{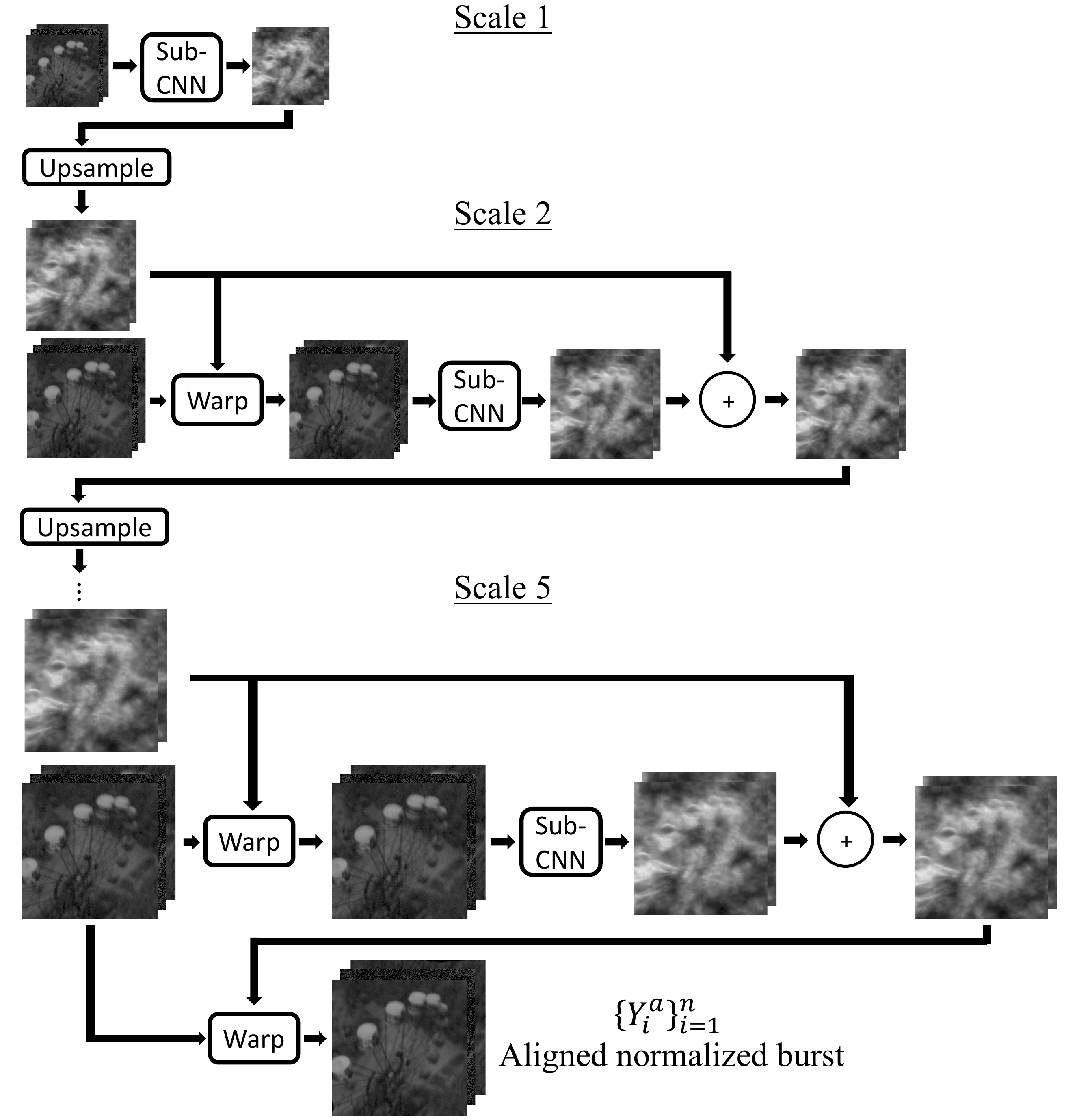}
\end{center}
   \caption{Our flow network obeys the architecture proposed by Kalantari and Ramamoorthi~\cite{kalantari2019deep}. Our sub-CNN comprises four $5\times5$ convolutions intertwined with ReLUs, with the number of channels being $n,100,50,25,2\left(n-1\right)$. In every scale, we feed the sub-CNN the appropriate pyramid level of the normalized burst, each frame in its own channel. We interpret the output as $XY$ displacement fields for each non-reference frame.}
\label{fig:flow}
\end{figure}

The goal of the flow network is to align the frames $\left\{Y_i\right\}_{i=1}^n$ according to a pre-selected reference one $Y_{i_0}$. Since the entire burst is captured during continuous motion, we choose $Y_{i_0}$ to be the middle frame, as we expect it to overlap the others significantly, making alignment easier. Aiming to deal with non-rigid motion and parallax, we use flow networks~\cite{dosovitskiy2015flownet,ilg2017flownet,ranjan2017optical,sun2018pwc}. However, these methods align individual images and therefore are prone to fail on degraded input. Kalantari and Ramamoorthi~\cite{kalantari2019deep} handle this concern by jointly aligning the reference's neighboring frames, thus complementing missing information. We follow the same architecture and simultaneously align the entire burst.

The network utilizes a hierarchical approach \cite{kalantari2019deep,ranjan2017optical,wang2017light}, which develops and refines a flow from a Gaussian pyramid of the input. At each level, from coarse to fine, we apply the same sub-CNN. At the coarsest level, we treat the sub-CNN's output as a displacement field, which we then upsample and apply to the next level before feeding it to the sub-CNN again. Then, we sum the output flow with the upsampled one to produce the next flow in the chain and repeat. This architecture, depicted in Fig. \ref{fig:flow}, has the benefit of producing large-scale and high-precision pixel displacements.

Note that since the exposure times are not stable, the burst's brightness varies from frame to frame and during training. Therefore, we normalize the frames according to their portion of the exposure time budget before feeding them into the flow network:
\begin{align}
    Y_{i}^{\mathrm{n}} = \frac{T}{\Delta t_i} Y_i.
    \label{eq:norm}
\end{align}

\subsection{Kernel prediction network}

\begin{figure*}[h]
\begin{center}
\includegraphics[width=0.95\linewidth]{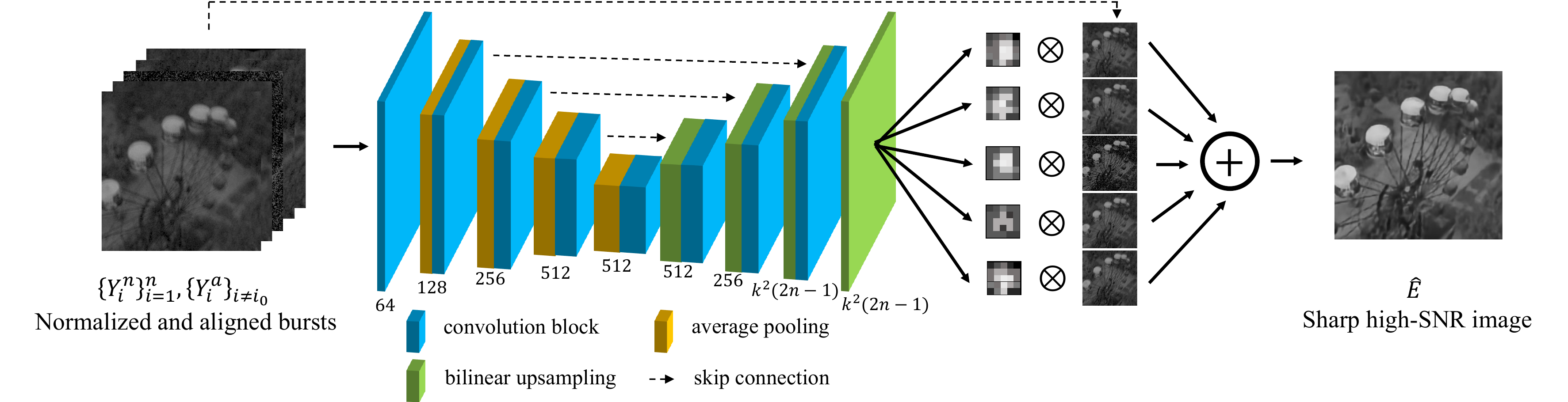}
\end{center}
   \caption{Our KPN follows the architecture of Mildenhall \etal~\cite{mildenhall2018burst}. All convolutions blocks are composed of three $3\times3$ convolutions intertwined with ReLUs. We feed the network both the original normalized burst and the aligned one, each frame in its own channel, and predict $k\times k$ kernels for each pixel in each given frame. We then apply the kernels to merge the frames into a single image.}
\label{fig:kpn}
\end{figure*}

\begin{figure}[h]
\begin{center}
\includegraphics[width=0.95\linewidth]{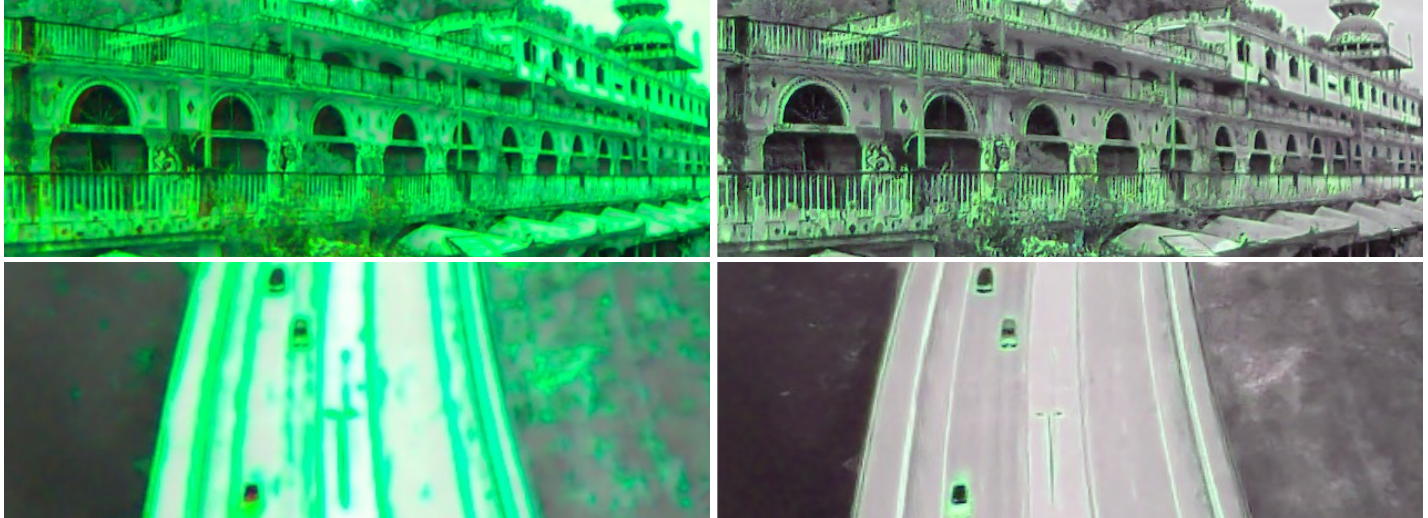}
\end{center}
       \caption{We reconstruct RGB images from the burst after applying the predicted kernels, for both KPN~\cite{mildenhall2018burst} (left) and our method (right). While KPN primarily uses the middle frame (corresponding to the color green), our algorithm manages to incorporate all frames. Since the middle frame’s learned exposure is lower than the others, it is mainly employed for restoring areas that suffer from more blur, such as edges and moving objects. The other two frames are averaged to denoise smoother surfaces or ones that are farther away from the camera and thus less sensitive to blur.}
\label{fig:kernels}
\end{figure}

Fig. \ref{fig:kpn} depicts our kernel prediction architecture, which follows the noise-blind version of the network suggested by Mildenhall \etal~\cite{mildenhall2018burst}. It is designed as an encoder-decoder with skip connections, which predicts adaptive kernels for each input pixel. The kernels are then applied to merge all given frames into the resulting image.

The use of pixel-adaptive kernels has a few key benefits. They reduce noise by pixel averaging without risking crossing edges and can also fix non-uniform blur. The latter is especially evident in our problem setting, where long focal lengths are commonly applied~\cite{whyte2012non}. Moreover, by picking the most reliable frames for each area in the final image, the kernels allow us to exploit the variability across the burst (see Fig. \ref{fig:kernels}). This choice is affected by each frame's exposure time and content but can also compensate for alignment artifacts by the flow network. Thus, similarly to Kalantari and Ramamoorthi~\cite{kalantari2019deep}, we merge the original and registered burst to allow the former to be picked over the other.

To generate the output, we first take the pixel-wise mean of the resulting frames:
\begin{align}
    \hat{E} = \Gamma\left(\frac{1}{2n-1} \left( \sum_{i=1}^n  Y_{i}^{\mathrm{n}} \otimes k_{i}^{\mathrm{n}} + \sum_{i \ne i_0} Y_{i}^{\mathrm{a}} \otimes k_{i}^{\mathrm{a}} \right)\right).
\end{align}
Here $\left\{Y_{i}^{\mathrm{a}}\right\}_{i=1,i\neq i_0}^n$ are the aligned frames, $\left\{k_{i}^{\mathrm{n}}\right\}_{i=1}^n$, $\left\{k_{i}^{\mathrm{a}}\right\}_{i\neq i_0}$ are the predicted location-dependent kernels, and $\otimes$ denotes the application of the kernel. Furthermore, to improve perceptual quality, we apply the differentiable gamma correcting function $\Gamma$~\cite{mildenhall2018burst}.

\section{Training}

\subsection{Loss function}

Given the ground-truth image $E$, we define our basic loss function as:
\begin{align}
    \ell_{\mathrm{basic}}\left(\hat{E},E\right) = \left\lVert\hat{E}-E\right\lVert_1 + \mu\left\lVert\nabla \hat{E}-\nabla E\right\lVert_1,
\end{align}
where $\nabla$ are the combined horizontal and vertical Sobel filters, and $\mu$ is a fixed constant. We choose the $\ell_1$ loss as it was shown to promote sharp outputs, while the second term is applied as a regularization aiming to suppress patterned artifacts~\cite{zhang2019deep}.

In our experiments, we found that while this loss yielded good results, the predicted kernels ignored $\left\{Y_{i}^{\mathrm{a}}\right\}_{i\neq i_0}$, as the flow network fails to properly align the burst. Thus, similarly to 
Mildenhall \etal~\cite{mildenhall2018burst}, we propose using an annealed loss term to encourage frame alignment. Unlike the mentioned authors, we do not apply the loss to each individual frame, since it might hinder the network from learning heterogeneous timing regimes. Instead, we sum each kernel entries to produce $1 \times 1$ kernels, and use them to merge the aligned burst:
\begin{align}
    \hat{E}^\mathrm{a} = \Gamma \left( \frac{\kappa}{\left(2n-1\right)\kappa_{\mathrm{a}}} \left( Y_{i_0}^\mathrm{n} \otimes K_{i_0}^\mathrm{n} + \sum_{i\neq0} Y_{i}^{\mathrm{a}} \otimes K_{i}^{\mathrm{a}} \right) \right).
    \label{eq:annealed}
\end{align}
Here $K$ denotes the sum over the corresponding indexed kernel, $\kappa$ is the pixel-wise sum over all $2n-1$ kernels, and $\kappa_{\mathrm{a}}$ is the pixel-wise sum over $\left\{K_{i}^{\mathrm{a}}\right\}_{i\neq i_0}$ and $K_{i_0}^{\mathrm{n}}$. Thus, our overall loss is
\begin{align}
    \ell\left(\hat{E},\hat{E}^\mathrm{a},E\right) = \ell_{\mathrm{basic}}\left(\hat{E},E\right) + \beta\alpha^t\ell_{\mathrm{basic}}\left(\hat{E}^{\mathrm{a}}, E\right),
\end{align}
where $\beta,\alpha\in\left(0,1\right)$ are fixed constant, and $t$ is the current iteration number. Hence, the loss in the first phase of training will produce gradients that compensate for the kernels' narrow receptive field by improving the flow network's performance.

\subsection{Synthetic dataset}

While standard burst and video denoising and deblurring algorithms operate on a few neighboring frames, our method must be evaluated on burst frames matching the irregular exposure regime dictated by our training on dense irradiance maps. Therefore, in the absence of any existing dataset that allows this flexibility, we generated our own.

Since our network is proposed as an alternative to mechanical gimbals, we create our data using 720p scenery drone videos from YouTube. We divide each video into short clips of 31 frames while skipping 500 frames between each clip. This process yielded around 10,000 different clips. Since undersampling may cause discontinuous artifacts in the motion trajectory, we follow other blur simulating works~\cite{brooks2019learning,nah2019ntire} and increase the frame rate using a CNN for frame interpolation~\cite{niklaus2017video2}. Although deep interpolation algorithms may create unreliable individual frames, they can still handle nonlinear motion and produce naturally looking blur when averaged. We feed each clip to the CNN eight times, resulting in $241$ frames. Since the middle frame is not artificially generated, we use it as the ground-truth.

At each training step, we randomly choose a $512\times512$ patch from the input clip. Although it already depicts camera motion, which is imperative for producing parallax and occlusions, we further augment the set by adding a controlled amount of motion using homographies. Since drone videos are usually filmed using long focal lengths, their primary source of blur is ego-rotations~\cite{whyte2012non}. We therefore apply randomly aggregated 3D rotations from frame to frame, which are computed using a pre-selected camera intrinsic matrix. Choosing the range of angles to rotate by affects the amount of blur in the final burst. However, these transformations may damage the original video quality and introduce black margins. To mitigate this degradation, we crop the patch around its center to generate a $256\times256$ clip and spatially downsample it using an anti-aliasing filter. This process produces $128\times128$ clips suffering from evident camera shake.

The final step is converting the clip to its corresponding irradiance values. To accomplish this, we first invert gamma correction, thus employing a linear color space, in which the pixel values are proportional to the mean number of incoming photons. Finally, we multiply the clip by a random scalar from a fixed range, which suits our target SNR.

\section{Experiments}

\subsection{Baseline}

We analyze the performance of our method against two traditional approaches: burst denoising and single image deblurring. While our algorithm incorporates non-uniformly-exposed bursts, which may suffer from both noise and blur, these alternatives often handle a single kind of degradation. Therefore, to allow a fair comparison, we apply burst denoising to a uniformly-exposed burst with the same overall budget and image deblurring to a single image captured using the budget in its entirety.

In our comparison, we use recent state-of-the-art deep learning algorithms. For image deblurring, we apply DMPHN~\cite{zhang2019deep2}, DeblurGAN-v2~\cite{kupyn2019deblurgan}, and the analysis-synthesis network pair~\cite{kaufman2020deblurring}. To conduct a fair comparison, we re-train one representative method on million full-exposure images generated from our forward model. As the latter method does not have a publicly available training code, we re-train DeblurGAN-v2, which showed better results than DMPHN. As for burst denoising, we re-train KPN~\cite{mildenhall2018burst} on our dataset.

\subsection{Synthetic images}

\begin{figure*}[h]
\begin{center}
\includegraphics[width=0.95\linewidth]{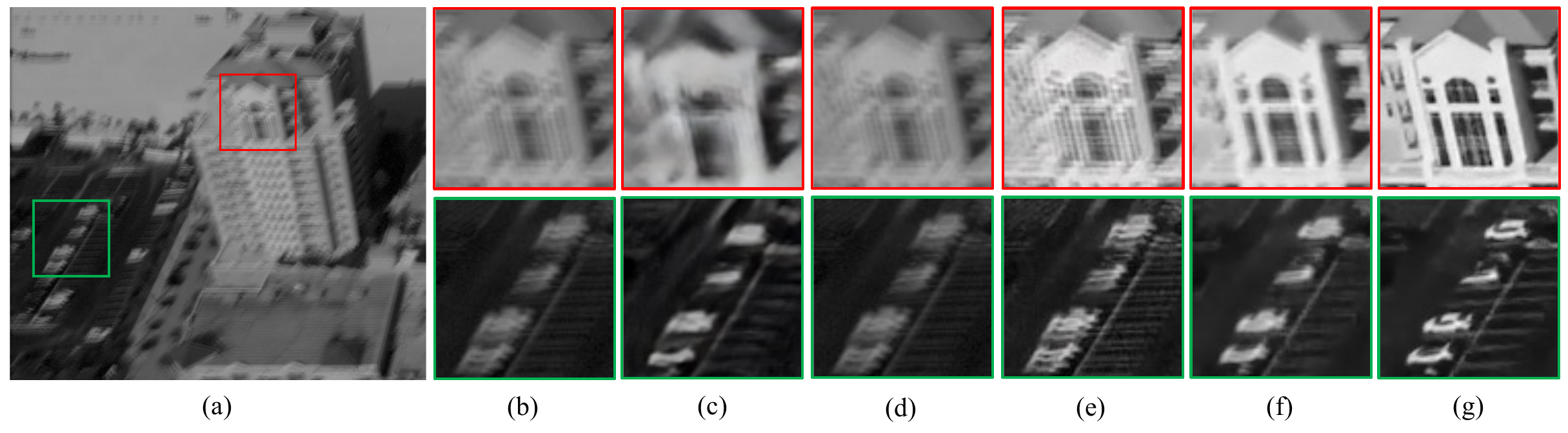}
\end{center}
   \caption{Example result from our synthetic test set. (a,b) Full-exposure image; Reconstruction using: (c) DMPHN~\cite{zhang2019deep2}; (d) Analysis-synthesis network pair~\cite{kaufman2020deblurring}; (e) DeblurGAN-v2~\cite{kupyn2019deblurgan}; (f) KPN~\cite{mildenhall2018burst}; Our approach (g). We showcase more results in the supplement.}
\label{fig:synthetic}
\end{figure*}

\begin{table}
\begin{center}
\begin{tabular}{|l|c|c|}
\hline
Method & PSNR & SSIM \\
\hline\hline
DMPHN~\cite{zhang2019deep2} & 17.99 & 0.6391 \\
Analysis-synthesis network pair~\cite{kaufman2020deblurring} & 18.6 & 0.7308 \\
DeblurGAN-v2~\cite{kupyn2019deblurgan} & 23.12 & 0.7230 \\
KPN~\cite{mildenhall2018burst} & 28.46 & 0.8296 \\
Ours & \bf{31.42} & \bf{0.8948} \\
\hline
\end{tabular}
\end{center}
\caption{Average PSNR and SSIM for our synthetic test set.}
\label{tab:average}
\end{table}

We first evaluate our method on a test set of 276 clips obtained using the same procedure as our training set, although originating from different videos. We present a quantitative comparison to our baseline in Table \ref{tab:average}. Our model outperforms traditional methods, obtaining an improvement of almost $3\si{\dB}$ over KPN.

Fig. \ref{fig:synthetic} presents an example result. As we can see, all three single image deblurring algorithms fail to deblur the image: DMPHN's output is distorted, while the other two exhibit ghosting artifacts. This is expected, as the full-exposure image demonstrates an evident amount of blur, which is hard to mitigate without any complementary information. Furthermore, this image also suffers from noise, raising the task difficulty even more.

On the other hand, multi-frame methods show better performance. KPN manages to clean the fixed-exposure bursts' noise and moderately improve the resulting images' sharpness. Nevertheless, due to the lack of frame diversity, it fails in capturing the fine details of the scene. Our approach of learning the burst exposures is the best alternative, producing a clean and sharp output.

\subsection{Real images}

\begin{figure*}[h]
\begin{center}
\includegraphics[width=0.95\linewidth]{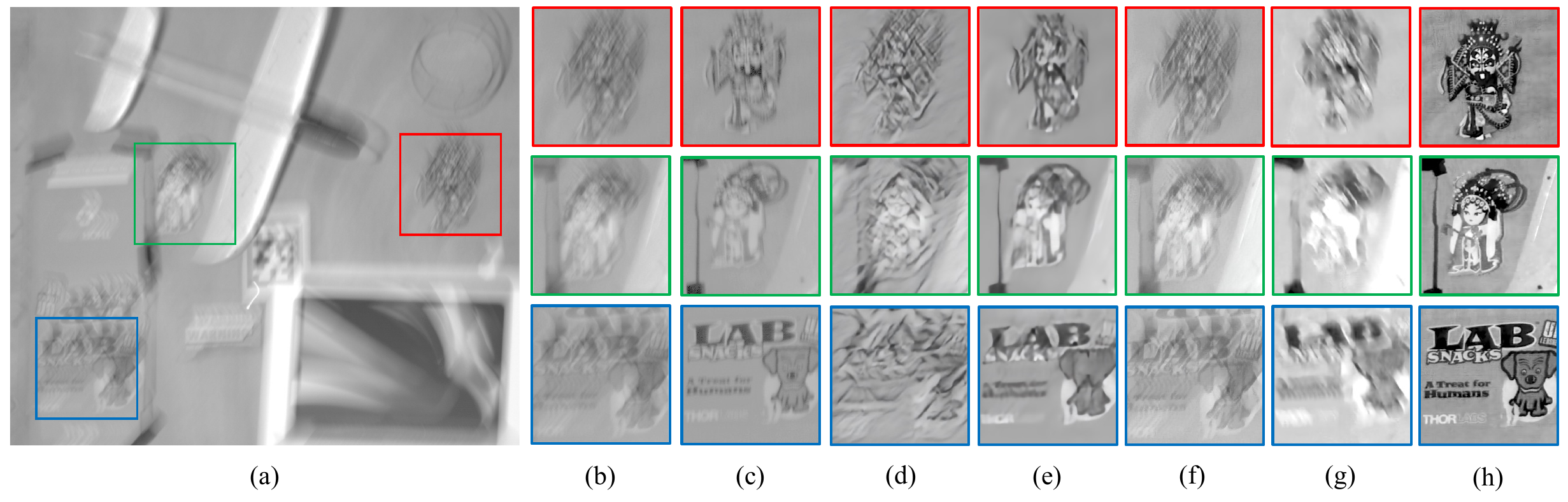}
\end{center}
   \caption{Example result on real images. (a,b) Full-exposure image; (c) Exemplary burst frame; Reconstruction using: (d) DMPHN~\cite{zhang2019deep2}; (e) Analysis-synthesis network pair~\cite{kaufman2020deblurring}; (f) DeblurGAN-v2~\cite{kupyn2019deblurgan}; (g) KPN~\cite{mildenhall2018burst}; Our approach (h).}
\label{fig:real_in}
\end{figure*}

We further evaluate our method on real images acquired using two different cameras at two different scenes: an indoor one with controlled light conditions and an outdoor one. (See the supplement for a full description of our setup.) Fig. \ref{fig:real_out} and \ref{fig:real_in} present a qualitative comparison, showing that the methods' perceptual quality appears to be generally consistent with the synthetic assessment, with our learned-exposure model significantly outperforming other approaches. However, this time, the analysis-synthesis network pair achieves the best results among the single image deblurring methods, despite not being re-trained for the task, while the re-trained DeblurGAN-v2 yields poor results. This deviation can be explained by the training SNR not precisely matching the de-facto one. Nevertheless, it proves our method's advantage even when the training conditions are not met.

Finally, we note that as the camera we used for the indoor experiment suffers from harsh fixed-pattern noise at low exposures, traces of such patterns can be found in the final output. However, this phenomenon was not replicated by the other camera, and we expect that incorporating non-uniformities in the forward model should mitigate it. We leave this revision for future work.

\subsection{Impact of noise and blur levels}

\begin{figure}[h]
\begin{center}
\includegraphics[width=1\linewidth]{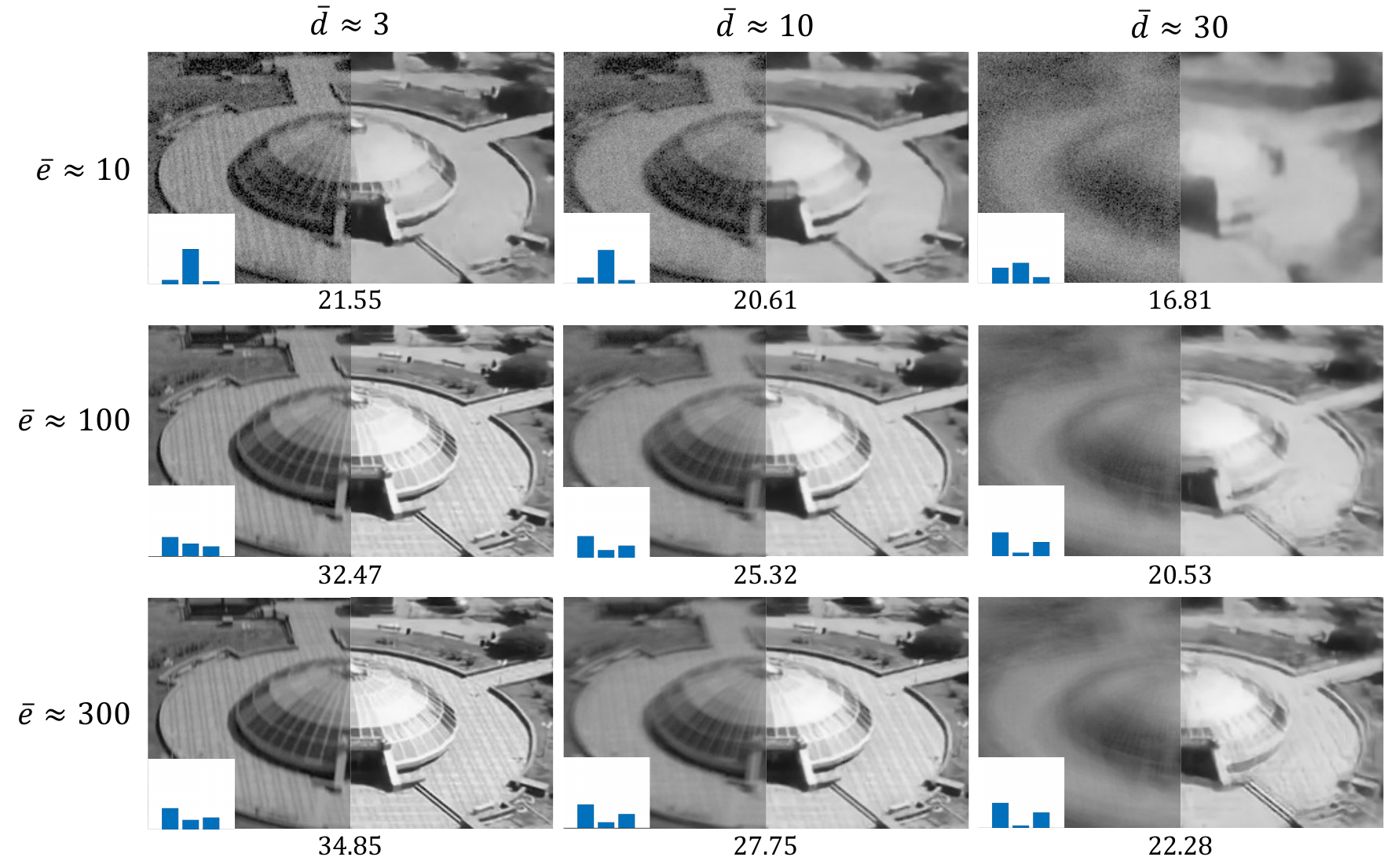}
\end{center}
   \caption{We compare our method's performance for different noise and blur conditions, displaying the full-exposure image (left), the network's output (right), its PSNR (under each image), and the learned exposures (bottom-left). Here $\bar{d}$ denotes the mean blur kernel diameter in pixels, and $\bar{e}$ is the mean number of photoelectrons, both estimated at full-exposure.}
\label{fig:rectangle}
\end{figure}

The main factors that determine the trade-off between deblurring and denoising are illumination and camera motion. Thus, they have a substantial effect on our method's learned exposures and reconstruction quality. Fig. \ref{fig:rectangle} explores these factors' impact by displaying the method's performance on the same synthetic scene, altered with varying noise and blur levels. To achieve optimal results, we re-train the model for each working point on suitable ranges of SNR and blur while keeping the rest of the hyper-parameters fixed, including the predicted kernel size, which we set to $5\times5$. (We list the full configuration in the supplement.)

Interestingly, the middle frame's exposure, which acts as the reference frame in the flow network, corresponds directly to the mentioned conditions, with it increasing as the SNR decreases and decreasing as camera motion increases, allowing the frame to be as sharp and clean as possible. The model's performance also varies accordingly. It yields visually-pleasing results, even when the blur kernel exceeds the boundaries of the predicted one. However, like the floor's checkered pattern, some details disappear in severe cases of noise and blur.

\section{Conclusions}

In this work, we presented a new approach for image stabilization via joint burst denoising and deblurring for fast unstabilized cameras. We further proposed an end-to-end learning scheme for optimizing the camera's exposure regime along with a reconstruction model, made possible via a novel differentiable layer simulating the camera sensor. This method's key benefit is its ability to exploit the trade-off between high SNR and strong blur at long exposure, and vice versa. Synthetic and real results suggest that this approach significantly improves current deep state-of-the-art methods, both perceptually and quantitatively.

\section{Acknowledgement}

This research was kindly supported by the Smart Imaging Consortium that has been financed by MAGNET program of the Israel Innovation Authority (IIA).

\emptyauthor
\title{Supplemental Material}
\maketitle
\appendix
\thispagestyle{empty}

\section{Backpropagation through the sensor forward model}

\subsection{Noise generation} 

Direct differentiation of (\ref{eq:shot}) is impossible, as it involves sampling from a parametrized distribution of a discrete random variable. Therefore, we resort to estimating the gradients of the expectation over the loss. This is commonly achieved using one of two methods: the score function~\cite{henderson2006handbooks} or  reparametrization~\cite{kingma2013auto}. Since score methods deviate from the orthodox backpropagation procedure~\cite{schulman2015gradient}, we opted for the latter.
    
Recently, Joo \etal~\cite{joo2020generalized} introduced the Generalized Gumbel-Softmax (GenGS) reparametrization, which can approximate any discrete, non-negative, and finite-mean random variable. They achieve this by truncating its support to a finite number of bins and relaxing the resulting categorical distribution into a continuous form using the Gumbel-Softmax reparametrization~\cite{jang2016categorical}. Nevertheless, this method has a few drawbacks which need to be addressed.
    
First, it requires setting two hyperparameters: the temperature $\tau_\mathrm{GenGS}$, which controls the smoothness of the resulting distribution, and the number of bins in its categorical support $n_\mathrm{GenGS}$. As $\tau_\mathrm{GenGS}$ approaches zero, sampling becomes increasingly discrete, while the gradient variance grows. Therefore, we apply the approach of Jang \etal~\cite{jang2016categorical} by starting with a high temperature and annealing it towards zero along training.
    
Choosing $n_\mathrm{GenGS}$ is less trivial. As the support of any Poisson distribution is infinite, increasing this hyperparameter will result in a better categorical approximation. However, since the latter are represented using one-hot vectors, memory complexity will linearly increase as well. This is especially problematic in our case, where we simultaneously sample from a different Poisson distribution for each pixel. To solve this issue, we rely on a corollary of the central limit theorem, which indicates that for high mean rate of arrivals (\eg more than 1000~\cite{mendenhall2016statistics}), a Poisson distribution may be approximated by a Gaussian one. Therefore, we use GenGS for low photon counts, and a Gaussian distribution for larger ones. This allows setting $n_\mathrm{GenGS}$ to a relatively low value.

We observed that the above procedure increases performance by around $0.1\si{\dB}$ for low to moderate SNR levels, compared to a simple Gaussian approximation.

\subsection{Quantization}

Lastly, we note that the derivative of the rounding operator in (\ref{eq:quantization}) is a.e. zero. Upon first inspection, this disallows backpropagation through the quantization step in (\ref{eq:dn}). However, if we consider the noisy additive perturbations to the quantizer input in (\ref{eq:total}), we may treat this process as stochastic and use the derivative of the expectation of the quantized value, which is smooth. We approximate this quantity by the identity straight through estimator~\cite{bengio2013estimating}.

\section{Technical details}

\subsection{Implementation details}

We implement our model using PyTorch~\cite{paszke2019pytorch},  Kornia~\cite{riba2020kornia} and PyTorch3D~\cite{ravi2020accelerating}, and train it for one million iterations on 4 NVIDIA GeForce RTX 2080 Ti GPUs. Since we work with 241 frames for each training example, we use NVIDIA DALI\footnote{https://github.com/NVIDIA/DALI} to accelerate loading times and upload batches of 2 clips directly to each GPU. We optimize using the ADAM solver~\cite{kingma2014adam} with a learning rate of $10^{-4}$. Training takes roughly 1-2 days.

We produce bursts comprised of $n=3$ frames. The forward model was configured with $\tau_1$ being 90\% of the full-well capacity of the respective camera. GenGS was applied for less than $1000$ incoming electrons with $n_{\mathrm{GenGS}}=1200$. We set $\tau_{\mathrm{GenGS}}$ to $e^{-10^{-5}t}$, where $t$ is the iteration number, until a minimum of $0.1$ is reached. The predicted kernel size is $5\times5$. Our loss hyperparameters are $\mu=1,\alpha=0.9999886,\beta=100$. For a 10-bit 720p burst, evaluation takes approximately 0.7 seconds on a single GeForce RTX 2080 Ti GPU while requiring 7110MB of memory.

\subsection{Experiment setup}

\begin{table*}
\begin{center}
\begin{tabular}{|l|c|c|c|c|c|c|}
\hline
Experiment & Camera & $T$ & $\Delta t_{\mathrm{min}}$ & $\Delta t_{\mathrm{ro}}$ & Added blank slot ($\Delta t_4$) & Learned exposures \\
\hline\hline
Synthetic & KAYA Instruments JetCam19M & 3\si{\milli\second} &  0\si{\micro\second} & 500\si{\micro\second} & \ding{51} & 872, 234, 583\si{\micro\second} \\
Indoor & KAYA Instruments JetCam19M & 5\si{\milli\second} &  500\si{\micro\second} & 400\si{\micro\second} & \ding{55} & 1228, 503, 828\si{\micro\second} \\
Outdoor & FLIR BFS-U3-16S2M & 5\si{\milli\second} & 4\si{\micro\second} & 400\si{\micro\second} & \ding{55} & 1203, 347, 801\si{\micro\second} \\
\hline
\end{tabular}
\end{center}
\caption{Experiment configurations.}
\label{tab:setup}
\end{table*}

We list the selected camera and exposure configuration of each conducted experiment in Table \ref{tab:setup}.

\paragraph{Indoor experiment.}

To enable different vibration modes with multiple degrees of freedom during acquisition, we mounted a Turnigy MultiStar $570KV$ drone motor with electronic speed control asymmetrically on top of the camera. We set the camera indoors, five meters in front of a wall depicting several printed signs. We illuminated it by a non-flickering $3000\si{\kelvin}$ led measured at around $50$-$80\si{\lux}$ from the camera's viewpoint. We used a $50\si{\milli\meter}$ fixed focal lens with an f-number of $1.8$. To mitigate fixed-pattern noise (FPN), we applied dark frame subtraction and flat-field correction as provided by the camera's ISP. Since we manually captured the burst, the blank interval between consecutive frames was not fixed in practice.

\paragraph{Outdoor experiment.}

\begin{figure}
\begin{center}
\includegraphics[width=0.72\linewidth]{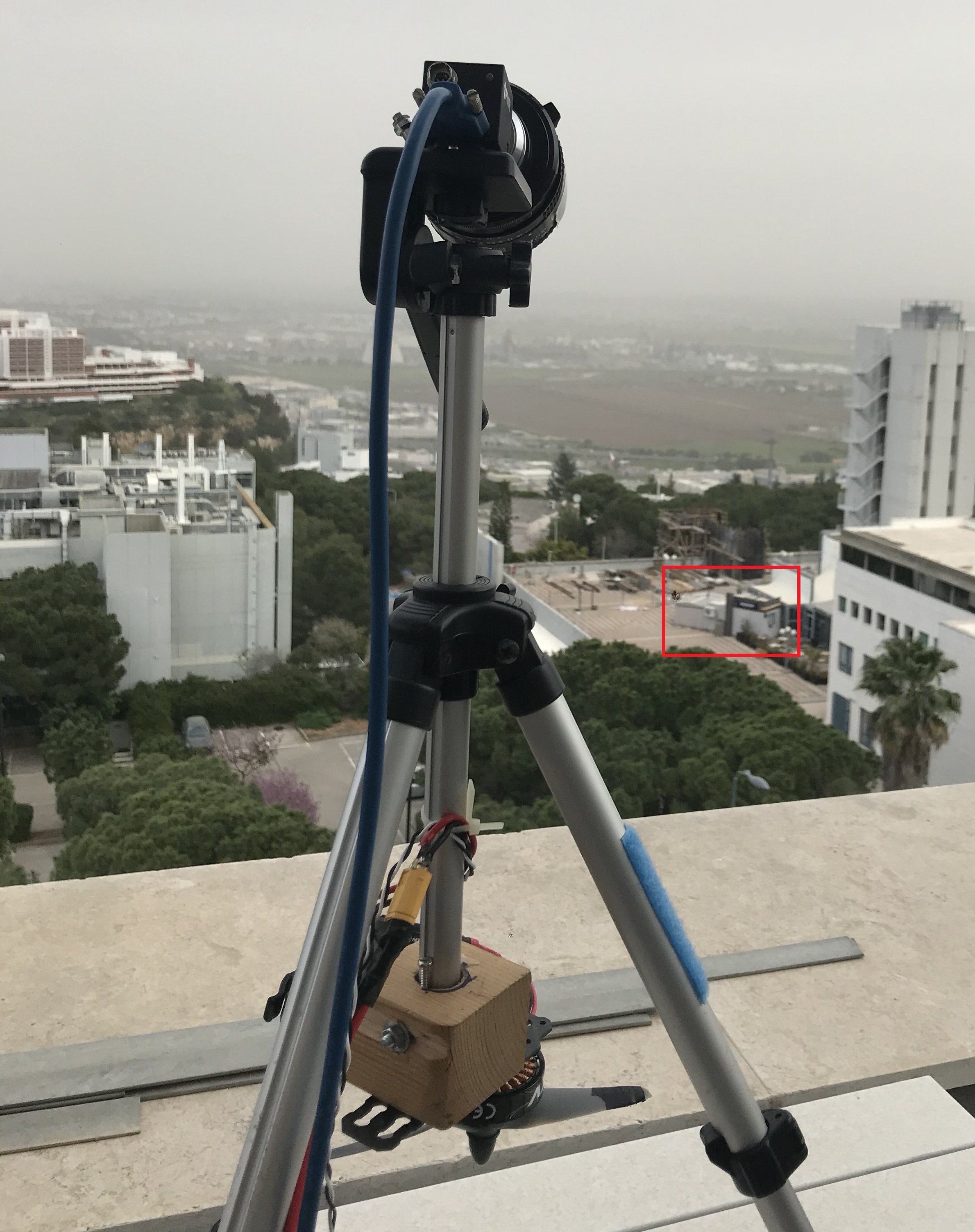}
\end{center}
   \caption{Outdoor experiment setup. The acquired scene is marked in red.}
\label{fig:Experiemnt setup}
\end{figure}

We chose FLIR Blackfly S USB 3.1 as a low-cost camera with a high frame rate and a flexible interface. The camera offers $226$ FPS at a resolution of $1440\times1080$ pixels and a configurable array of up to $8$ exposures, allowing the subsequent capture of the desired learned-exposure burst, corresponding fixed-exposure burst, and full-exposure image for comparison. We chose a lens with a focal length of $50\si{\milli\meter}$ and set the f-number to $22$.

We set the camera on a tripod and attached the same motor from the previous experiment underneath it to create vibrations, as depicted in Fig. \ref{fig:Experiemnt setup}. We supplied the motor with a voltage of $13.8\si{\volt}$, resulting in approximately $7850$ RPM. Finally, we increased the vibration amplitude by unbalancing the rotor blades.

\section{Additional results}

\subsection{Additional synthetic results}

\begin{figure*}
\begin{center}
\includegraphics[width=1\linewidth]{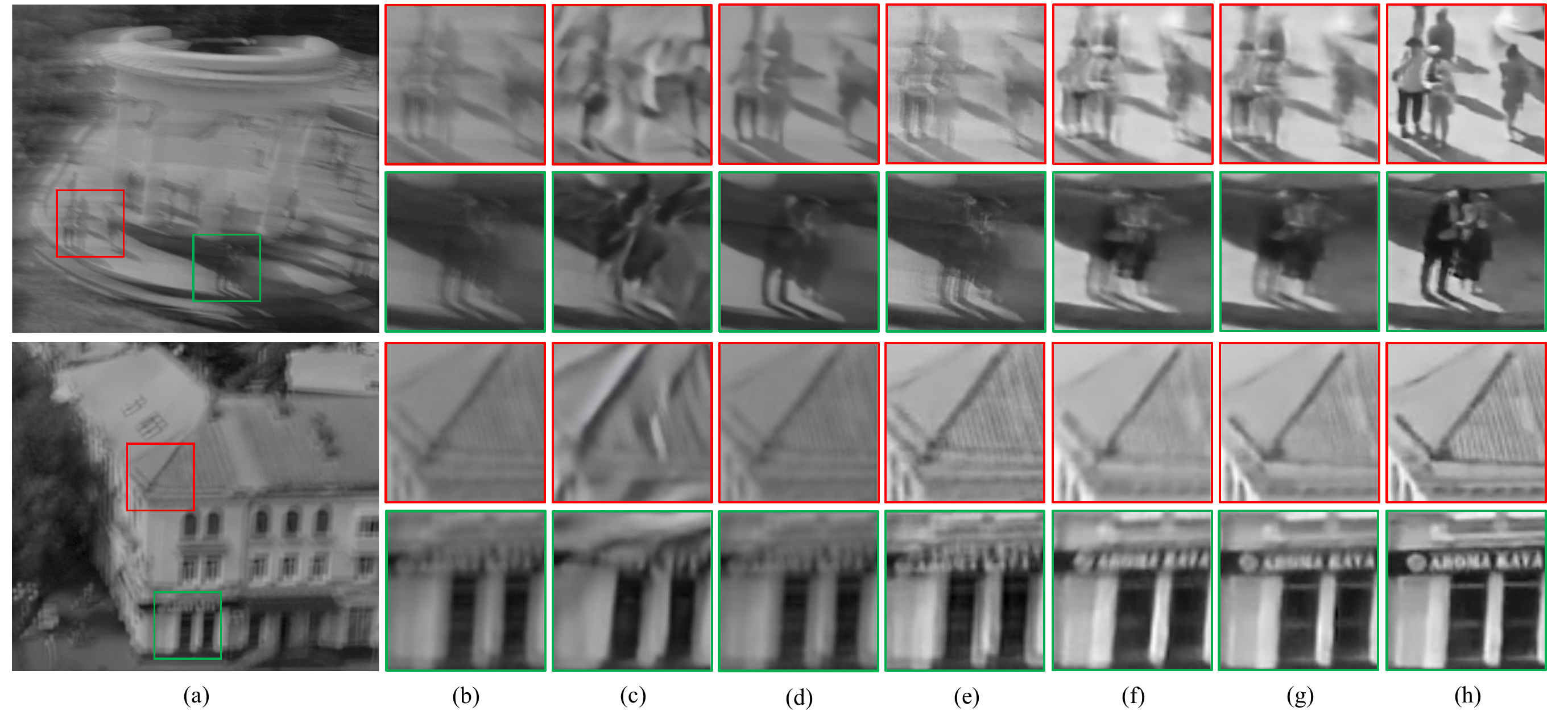}
\end{center}
   \caption{Example results from our synthetic test set. (a,b) Full-exposure image; Reconstruction using: (c) DMPHN~\cite{zhang2019deep2}; (d) Analysis-synthesis networks pair~\cite{kaufman2020deblurring}; (e) DeblurGAN-v2~\cite{kupyn2019deblurgan}; (f) KPN~\cite{mildenhall2018burst}; Our approach using (g) fixed uniform, and (h) learned exposures.}
\label{fig:more_synthetic}
\end{figure*}

We include more results on our synthetic test set in Fig. \ref{fig:more_synthetic}.

\subsection{Ablation study}

\begin{table}
\begin{center}
\begin{tabular}{|l|c|c|}
\hline
Missing component & PSNR \\
\hline\hline
Exposure learning & 28.72 \\
Flow network & 30.62 \\
Exposure normalization & 30.96 \\
Annealed loss term & 31.18 \\
None & \bf{31.42} \\
\hline
\end{tabular}
\end{center}
\caption{Average PSNR of ablated models.}
\label{tab:ablation}
\end{table}

\begin{figure}
\begin{center}
\includegraphics[width=1\linewidth]{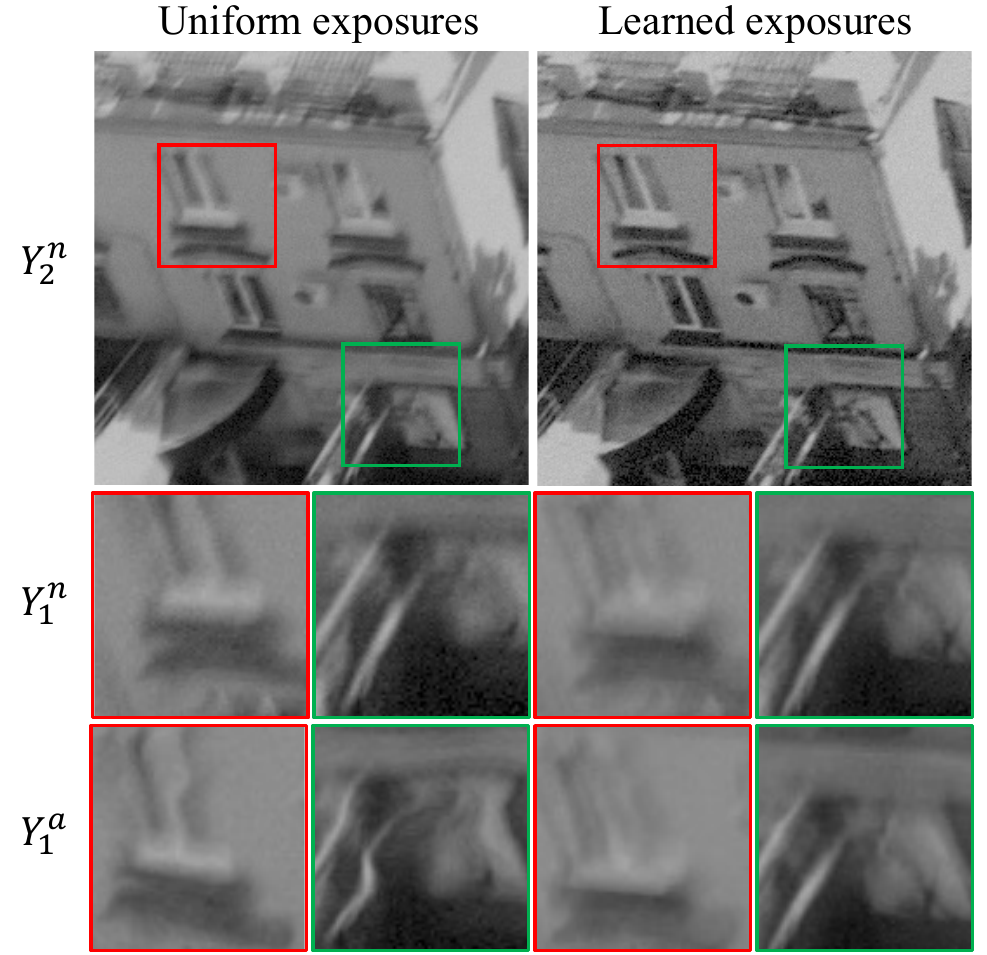}
\end{center}
   \caption{We compare the flow network's performance when using uniform and learned exposures. Since the uniform burst has a blurry reference frame, alignment results in evident artifacts. On the other hand, learning the burst's exposures enables proper alignment.}
\label{fig:registration}
\end{figure}

We conduct an ablation study to assess the contribution of each network module to our proposed pipeline. Table \ref{tab:ablation} presents the average PSNR obtained on our synthetic test set by different ablated models. As we can see, the most impactful component is the forward model, which enables learning optimal exposures and yields a significant performance gain of $2.7\si{\dB}$. This improvement is further demonstrated by comparing our model's uniform-exposure output to the learned-exposure one in Fig. \ref{fig:more_synthetic}. Moreover, pre-aligning the frames via the flow network improves results by $0.8\si{\dB}$. It is also evident that exposure normalization (\ref{eq:norm}) and our annealed loss term (\ref{eq:annealed}) are critical to the proper convergence of the learning process.

Careful examination of the flow network's performance further reveals the importance of using learned exposures besides forcing frame diversity. Fig. \ref{fig:registration} presents the reference frame and an aligned neighboring frame in both the uniform- and learned-exposure regime. As we can see, while the reference frame in the uniform case suffers from noticeable blur, its adaptive counterpart is much sharper due to its shorter exposure. This difference gives the latter model an advantage over the other, which depicts severe alignment artifacts. This observation demonstrates that exposure learning is not only beneficial for our end-goal reconstruction task but can also contribute to the performance of intermediate layers.

{\small
\bibliographystyle{ieee_fullname}
\bibliography{egbib}
}

\end{document}